\pdfoutput=1

\documentclass[11pt]{article}

\usepackage{EMNLP2022}

\usepackage{times}
\usepackage{latexsym}

\usepackage[T1]{fontenc}

\usepackage[utf8]{inputenc}

\usepackage{microtype}

\usepackage{inconsolata}
\usepackage{microtype}
\usepackage{subfigure}
\usepackage{amsmath}
\usepackage{tabularx}
\usepackage{graphicx}
\usepackage{booktabs}
\usepackage[linesnumbered,ruled,vlined]{algorithm2e}
  
\usepackage{CJK}

\title{Counterfactual Data Augmentation via Perspective Transition \\for Open-Domain Dialogues}

\author{Jiao Ou\textsuperscript{\rm 1,2}, \textbf{Jinchao Zhang\textsuperscript{\rm 3}}, Yang Feng\textsuperscript{\rm 1,2}\thanks{$^*$Work done while Jiao Ou was interning at WeChat AI. Yang Feng is the corresponding author. Our code is public at https://github.com/ictnlp/CAPT.}, \textbf{Jie Zhou\textsuperscript{\rm 3}}\\ 
\textsuperscript{\rm 1} Key Laboratory of Intelligent Information Processing,\\ Institute of Computing Technology, Chinese Academy of Sciences (ICT/CAS)\\
\textsuperscript{\rm 2} University of Chinese Academy of Sciences\\
\textsuperscript{\rm 3} Pattern Recognition Center, WeChat AI, Tencent Inc, China\\
{ \{oujiao17b,fengyang\}@ict.ac.cn, \indent \{dayerzhang,withtomzhou\}@tencent.com}
}

\begin{document}
\maketitle
\begin{abstract}
The construction of open-domain dialogue systems requires high-quality dialogue datasets.
The dialogue data admits a wide variety of responses for a given dialogue history, especially responses with different semantics.
However, collecting high-quality such a dataset in most scenarios is labor-intensive and time-consuming.
In this paper, we propose a data augmentation method to automatically augment high-quality responses with different semantics by counterfactual inference.
Specifically, given an observed dialogue, our counterfactual generation model first infers semantically different responses by replacing the observed reply perspective with substituted ones.  
Furthermore, our data selection method filters out detrimental augmented responses.
Experimental results show that our data augmentation method can augment high-quality responses with different semantics for a given dialogue history, and can outperform competitive baselines on multiple downstream tasks.
\end{abstract}

\section{Introduction}


Open-domain dialogue systems have attracted much attention~\cite{10.1145/3166054.3166058,DBLP:journals/tois/HuangZG20,ni2021recent,FU202214} due to their potential applications.
Generally, training open-domain dialogue systems requires high-quality dialogue datasets.
The dialogue data admits a wide variety of responses for a given dialogue history~\cite{hou-etal-2018-sequence}.
Specifically, a given dialogue history can exist many valid responses with different semantics, and the response of each semantic information can also have abundant alternative expressions~\cite{Li_Qiu_Tang_Chen_Zhao_Yan_2019}.
However, manually collecting high-quality such datasets is usually labor-intensive and time-consuming in practice.


A feasible solution to address this problem is to use data augmentation techniques.
Currently, some data augmentation methods have been used in open-domain dialogues~\cite{sennrich-etal-2016-improving,niu-bansal-2019-automatically,Li_Qiu_Tang_Chen_Zhao_Yan_2019,cai-etal-2020-data,zhang-etal-2020-dialogue,xie2022targetside} to augment data. 
However, the augmented data have limited semantic differences from the observed data based on the restrained changes.
These existing methods only consider word- or sentence-level alternative expressions of the observed data without augmenting more valid responses with different semantics.   

\begin{figure}
    \centering 
    \includegraphics[width=0.95\linewidth]{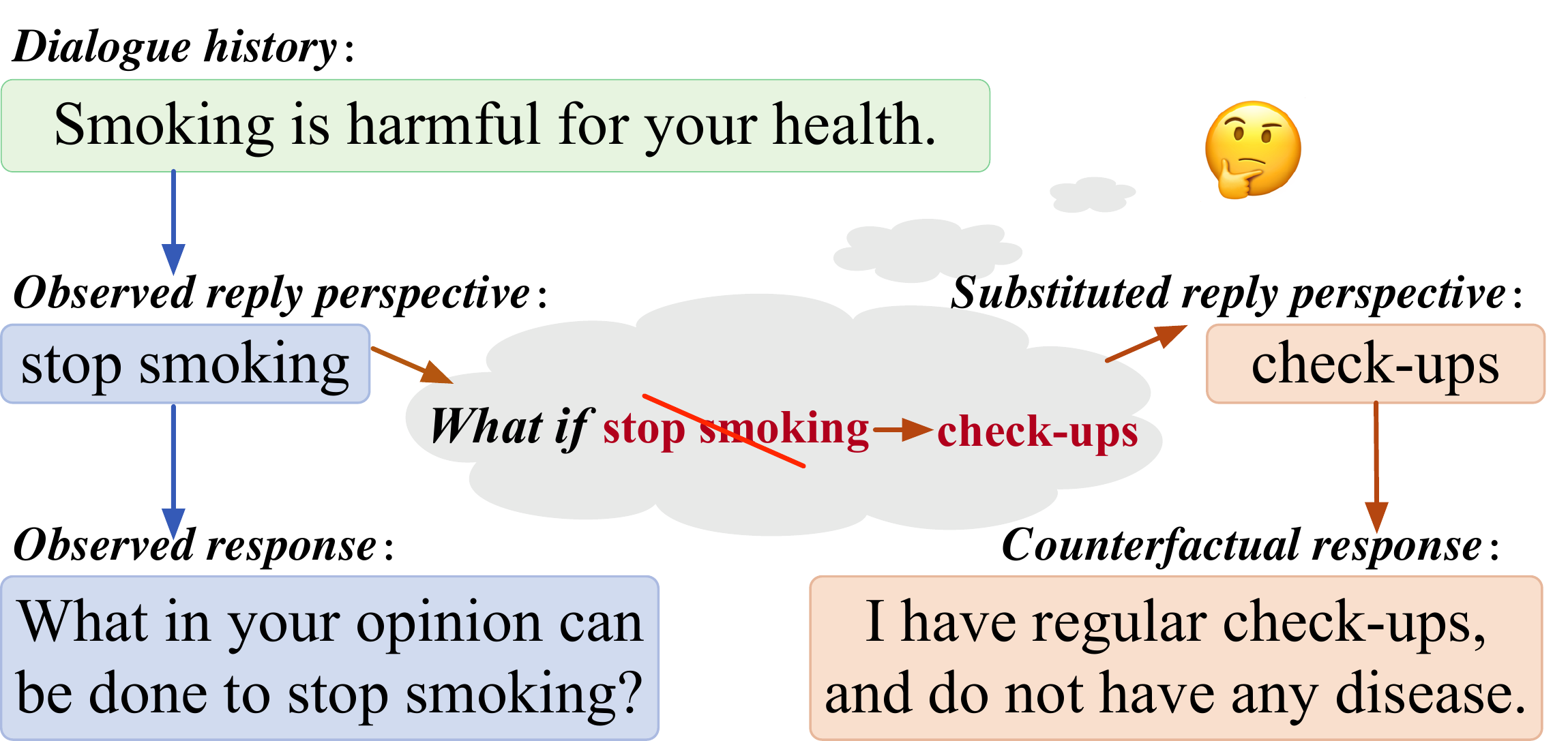}     
    \caption{An example of a counterfactual response, which is a semantically different response re-inferred by changing the observed reply perspective.
    } 
    \label{fig:intro} 
\end{figure}
In this paper, we propose to augment valid responses with different semantics for a given dialogue history.
Imagine that when humans infer different-semantic responses, they may naturally ask a question: Given an observed dialogue, what the response would happen if we change the \emph{reply perspective}, while keeping the current environment unchanged?
Answering this question will infer a different response, given an example shown in Figure~\ref{fig:intro}.
The imagination of different responses under the current environment is so-called \emph{counterfactual inference}~\cite{pearl2000models}, which
ensures the quality of inferred responses~\cite{zhu-etal-2020-counterfactual}.

Motivated by this, we propose a \textbf{C}ounterfactual data \textbf{A}ugmentation method via \textbf{P}erspective \textbf{T}ransition, CAPT for short, to generate counterfactual responses for a given observed dialogue.
CAPT interprets a counterfactual generation model as a structural causal model (SCM), which describes the generation process under the current environment.
The current environment is modeled by unobserved variables in the SCM that capture all unobserved but relevant factors that affect response generation.
Counterfactual responses are then generated by 
intervening in the reply perspective in the SCM, i.e., replacing the observed reply perspective with valid alternatives, while keeping these unobserved variables unchanged. 
To achieve an alternative, we first construct a shift graph based on all observed dialogues, which explicitly represents the shift associations between both focuses of attention on dialogue histories and their corresponding responses respectively.
We then randomly choose a focus on the given dialogue history and regard its 1-hop neighbors in the shift graph as candidates. 
A valid alternative can be predicted from these candidates.
After achieving all counterfactual augmented responses, the augmented data are further filtered using a data selection module.
Finally, we merge the observed data with this augmented data as training data for downstream tasks.

Experiment results indicate that CAPT can augment high-quality responses with different semantics, and our augmented data contributes to the performance improvement of both retrieval-based and generation-based open-domain dialogue models. 
Our contributions are summarized as follows:
(1) We propose a counterfactual data augmentation method via perspective transition to augment responses with different semantics for a given dialogue history. To the best of our knowledge, this is the first study to augment more valid responses with different semantics in open-domain dialogues.
(2) Automatic and manual evaluation show that CAPT generates semantically different responses, which can be further used to improve the performance of downstream tasks.
(3) Extensive experiments show that providing more responses with different semantics can further improve performance.

\section{Background}
In this section, 
we describe task definitions and review the concept of the structural causal model. 
Please see task definitions in Appendix~\ref{app:definition}.

\subsection{Structural Causal Model}
\paragraph{Definition.}
A structural causal model (SCM) consists a set of observed variables $\mathbf{V} = \{\boldsymbol{V}_1, \dots, \boldsymbol{V}_m\}$ and a set of independent unobserved random variables $\mathbf{U} = \{\boldsymbol{U}_1, \dots, \boldsymbol{U}_m\}$ with distribution $P(\mathbf{U})$, which are connected by a set of functions $\mathbf{F} = \{f_1, \dots, f_m\}$.
Specifically, $\forall i$, $\boldsymbol{V}_i$ is caused by a set of parent variables $\mathbf{PA}_i$ and $\boldsymbol{U}_i$, i.e., $\boldsymbol{V}_i = f_i(\mathbf{PA}_i, \boldsymbol{U}_i)$, where $\mathbf{PA}_i \subseteq \mathbf{V} \setminus \boldsymbol{V}_i$ in the causal DAG~\cite{DBLP:conf/iclr/BuesingWZHRGL19}.

For the counterfactual generation model, it can be cast as an SCM with three observed variables, including \emph{dialogue history} $\boldsymbol{X}$ , \emph{reply perspective} $\boldsymbol{Z}$ and \emph{response} $\boldsymbol{Y}$.
The counterfactual generation SCM turns the conditional distribution $P(\boldsymbol{Y}|\boldsymbol{X}, \boldsymbol{Z})$ into a deterministic function $\boldsymbol{Y} = f(\boldsymbol{X}, \boldsymbol{Z}, \boldsymbol{U})$,
where $\boldsymbol{U}$ captures all unobserved but influential factors of the current environment, such as speaking style. 
The function $f$ is defined by the learned counterfactual generation model.
Overall, SCM can infer counterfactual responses given the known function $f$ and the posterior of the unobserved variable $P(\boldsymbol{U}|\boldsymbol{X}=\boldsymbol{x}, \boldsymbol{Z}=\boldsymbol{z}, \boldsymbol{Y}=\boldsymbol{y})$.

\begin{figure*}
    \centering 
    \includegraphics[width=0.9\linewidth]{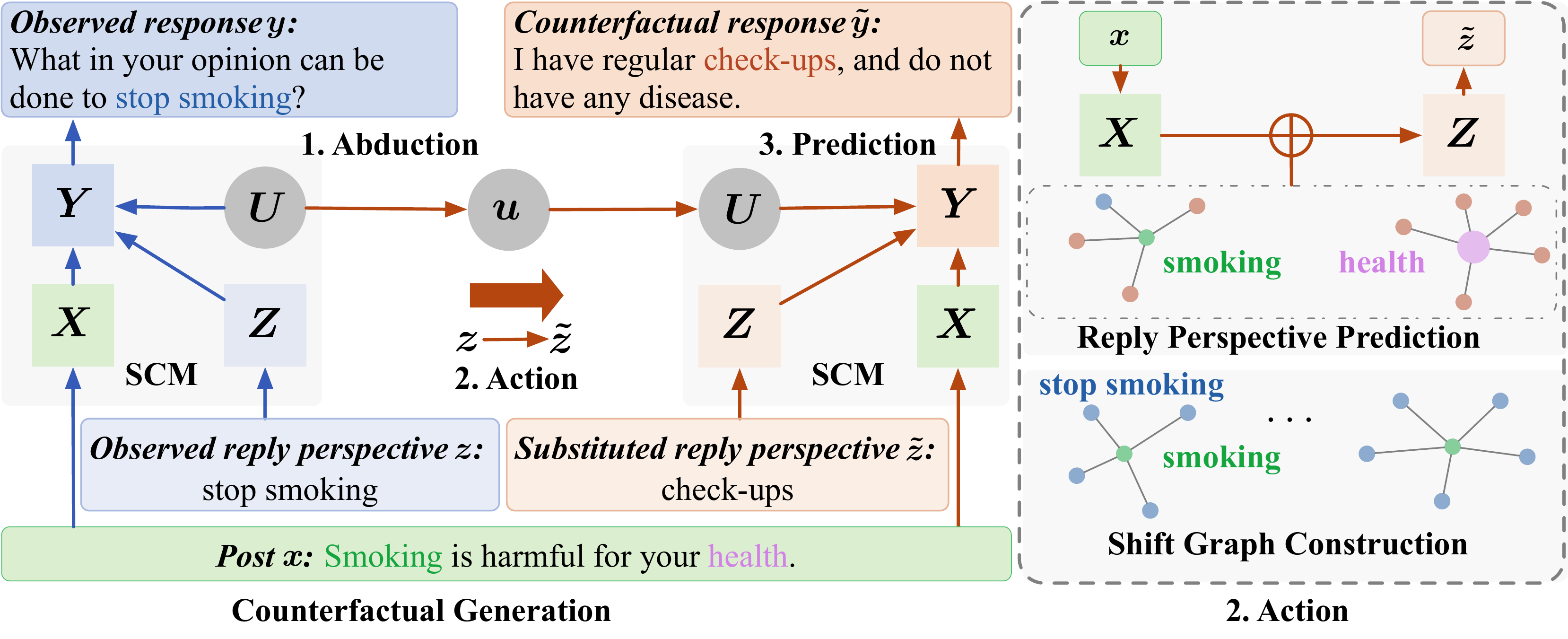}     
    \caption{The three-step procedure of counterfactual generation: (1) Abduction: we estimate the ``current environment of the SCM'' $\boldsymbol{u}$ where the observed response $\boldsymbol{y}$ occurs. 
    (2) Action: we perform an intervention on $\boldsymbol{Z}$ in the SCM by replacing $\boldsymbol{z}$ with $\boldsymbol{\tilde{z}}$. For obtaining $\boldsymbol{\tilde{z}}$, we first construct the shift graph by characterizing each observed shift between the focus (e.g., smoking) on $\boldsymbol{x}$ and the reply perspective $\boldsymbol{z}$ (e.g., stop smoking) in $\mathcal{D}$. 
    We then randomly choose a focus on $\boldsymbol{x}$ , e.g. health, and regard its 1-hop neighbors as candidates. Finally, we predict $\boldsymbol{\tilde{z}}$ from these candidates conditioned on the chosen focus of $\boldsymbol{x}$ and the post $\boldsymbol{x}$.
    (3) Prediction: the counterfactual response is generated based on the post $\boldsymbol{x}$ and the alternative $\boldsymbol{\tilde{z}}$ from the inferred $\boldsymbol{u}$.
    } 
    \label{fig:method} 
\end{figure*}

\paragraph{Intervention.}
Before observing what the observed variable $\boldsymbol{V}_i$ would happen, an intervention would be given on a parent variable $\boldsymbol{V}_j$, $\boldsymbol{V}_j \in \mathbf{PA}_i$, where the intervention in the SCM is an action by changing the observed value.
For the counterfactual generation SCM, the intervention is to replace the observed value $\boldsymbol{z}$ of the \emph{reply perspective} $\boldsymbol{Z}$ with a different value $\boldsymbol{\tilde{z}}$.

\paragraph{Counterfactual Inference.}
Given an SCM and observed a variable $\boldsymbol{V}_i = \boldsymbol{v}_i$,
counterfactual inference answers the question that what the observed variable $\boldsymbol{V}_i$ would have changed if a parent variable $\boldsymbol{V}_j$ has been intervened while keeping the current environment unchanged.
Accordingly, generating a counterfactual response involves a query about what the response $\boldsymbol{Y}$ would have happened if an intervention is taken by setting $\boldsymbol{Z}$ as a different value $\boldsymbol{\tilde{z}}$, rather than the observed value $\boldsymbol{z}$.

Overall, to generate counterfactual responses, we can follow a three-step procedure~\cite{pearlcausal}:
(1) \textbf{Abduction}: Predict the ``current environment of the SCM'' , i.e., compute the posterior $P(\boldsymbol{U}|\boldsymbol{X}=\boldsymbol{x}, \boldsymbol{Z}=\boldsymbol{z}, \boldsymbol{Y}=\boldsymbol{y})$ and sample $\boldsymbol{u}$ from it.
(2) \textbf{Action}: Perform an intervention by replacing the observed value $\boldsymbol{z}$ of $\boldsymbol{Z}$ with a different value $\boldsymbol{\tilde{z}}$. 
(3) \textbf{Prediction}: Reason a counterfactual response $\boldsymbol{\tilde{y}}$, 
given the posterior sample $\boldsymbol{u}$ and the known function $f$.


\section{Method}
In this section, our goal is to take an input dialogue sample $(\boldsymbol{x}, \boldsymbol{y})$ and augment high-quality responses 
that have different semantics from $\boldsymbol{y}$.
To this end, in Section~\ref{method:generation}, we introduce a technique called \emph{Counterfactual Generation via Perspective Transition} for intervening in the observed reply perspective to augment responses under the current environment.
In Section~\ref{method:training}, we describe how to train those models involved in Section~\ref{method:generation}, including the reply perspective predictor and the counterfactual generator.
In Section~\ref{method:selection}, we design a data selection method, named \emph{Bi-directional Perplexity Selection}, to select high-quality augmented data.

\subsection{Counterfactual Generation via Perspective Transition}
\label{method:generation}
This paper mainly focuses on single-turn dialogues.
Given a post-response pair $(\boldsymbol{x}, \boldsymbol{y})$, we use the SCM to generate a counterfactual response $\boldsymbol{\tilde{y}}$ 
following the three-step procedure shown in Figure~\ref{fig:method}.

\paragraph{1. Abduction.}
This step is to estimate the unobserved variable given the observed sample $(\boldsymbol{x}, \boldsymbol{z}, \boldsymbol{y})$ (for more details about $\boldsymbol{z}$ see the action step).
Specifically, when generating the $t$-th token of $\boldsymbol{y}$, our counterfactual generator outputs a categorical distribution $P(Y_t|\boldsymbol{X}=\boldsymbol{x}, \boldsymbol{Z}=\boldsymbol{z}, \boldsymbol{Y}_{<t}=\boldsymbol{y}_{<t})$, where $\boldsymbol{y}_{<t}$ is the token sequence generated in the previous time step.
According to ~\citet{oberst2019counterfactual}, the impact of the unobserved random variable $\boldsymbol{U}_t$ is simulated by introducing Gumbel random noises.
Thus, we perform the Gumbel-Max Trick~\cite{Luce59} for this categorical distribution as follows,
\begin{equation}
\label{eq:abduction}
\begin{aligned}
    p_{tk} = & P(Y_t=k|\boldsymbol{X}=\boldsymbol{x}, \boldsymbol{Z}=\boldsymbol{z}, \boldsymbol{Y}_{<t}=\boldsymbol{y}_{<t}), \\
    y_t = & \underset{k=1, \ldots,|V|}{\arg \max}(\log p_{tk} + u_{tk}), 
\end{aligned}
\end{equation}
where $u_{tk} \sim \text{Gumbel}(0, 1)$ and $|V|$ denotes the vocabulary size.

Consequently, our counterfactual generation SCM transforms into a Gumbel-Max SCM~\cite{oberst2019counterfactual}.
The estimation of the unobserved variable is then to sample from the posterior distribution over these Gumbel random variables.
Fortunately, a straightforward way to infer posterior~\cite{NIPS2014_309fee4e} is utilizing the properties of the shifted Gumbel variables $g_{tk} = \log p_{tk}+ u_{tk}$: in the posterior, the maximum value is independent with the argmax of the shifted Gumbel variables and is distributed as a standard Gumbel.
Thus, we first let $y_t=k^*$ (* denotes the observed token) and sample the maximum value $g_{tk}^*$ from $\text{Gumbel}(0,1)$.
Secondly, we sample the remaining values $g_{tk}$ from the shifted Gumbel distribution $\text{Gumbel}(\log p_{tk}, 1)$ truncated at $g_{tk}^*$.
Then, for each index of $k$, a sample of $u_{tk}$ is obtained by subtracting off the location parameter $\log p_{tk}$ from $g_{tk}$.
Finally, the resulting sample $\boldsymbol{u}_t = [u_{t1}, \dots, u_{t|V|}]$ is used to infer the counterfactual responses.
\paragraph{2. Action.}
This step is to replace the observed reply perspective $\boldsymbol{z}$ with a substituted reply perspective $\boldsymbol{\tilde{z}}$. 
However, two sub-problems need to be addressed: \emph{representing the reply perspective} and \emph{predicting a substituted valid reply perspective}.
By observing human dialogues, we find that a reply perspective can be represented by a keyword, like ``stop smoking'' in Figure~\ref{fig:method}. 
It can be achieved based on the process that humans first naturally focus on a certain point of a given post like ``smoking'' and then would unconsciously shifting this focus point to another one.
The focus point of the post can be similarly represented by a keyword.
We name the focus point on the post and the shifted one as the \emph{focus} and \emph{reply perspective} respectively.
When humans have different focuses (e.g., ``health'' in Figure~\ref{fig:method}) or different shifts on the same focus, they will obtain substituted reply perspectives. 

To achieve valid alternatives, it is critical to make valid shifts from a focus.
We build a shift graph based on all observed samples, where head and tail vertices are focuses and reply perspectives respectively, and edges represent observed shifts between focuses and reply perspectives.
Inspired by ~\citet{xu-etal-2020-conversational} and ~\citet{zou-etal-2021-thinking},
we can regard 1-hop neighbors of a given focus as candidates and predict a valid alternative from these candidates.
It is based on the fact that the corresponding reply perspectives can be shared if posts containing the same focus have similar semantics.

We build the shift graph $\mathcal{G}$ with two steps: vertex construction and edge construction.
For vertex construction, 
we first exploit a rule-based keyword extraction method~\cite{campos2020yake} to identify salient keywords from utterances in the observed dialogue dataset $\mathcal{D}$.
To further identify the focus $\boldsymbol{c}$ from all keywords of $\boldsymbol{x}$, we use guidance from the future information (i.e., response) to select the keyword that is semantically closest to $\boldsymbol{y}$.
To identify the reply perspective $\boldsymbol{z}$, we select the keyword with the closest semantics to $\boldsymbol{c}$.
More concretely, we use cosine similarity between their embedding via BERT~\cite{devlin-etal-2019-bert} as the measure of semantic closeness, where each embedding is achieved by taking the average of the hidden state of each token.
For edge construction, we build an edge by connecting $\boldsymbol{c}$ with $\boldsymbol{z}$.
In this way, we characterize all shift associations in $\mathcal{D}$.

Once the shift graph is built, we predict $\boldsymbol{\tilde{z}}$ as 
\begin{equation}
    \boldsymbol{\tilde{z}} = \arg\max\nolimits_{\boldsymbol{\tilde{z}}} P(\boldsymbol{Z}|\boldsymbol{C}=\boldsymbol{\tilde{c}}, \boldsymbol{X}=\boldsymbol{x}, \boldsymbol{N}=\mathcal{N}(\boldsymbol{\tilde{c}})),
\end{equation}
which is given by a trained reply perspective predictor. 
Note that $\boldsymbol{\tilde{c}}$ can be any keyword in the post $\boldsymbol{x}$ and $\mathcal{N}(\boldsymbol{\tilde{c}}))$ denotes 1-hop neighbors of $\boldsymbol{\tilde{c}}$.

\paragraph{3. Prediction.}
This step is to generate the counterfactual response given the posterior sample $\boldsymbol{u}_t = [u_{t1}, \dots, u_{t|V|}]$.
Specifically, when generating the $t$-th token of the counterfactual response, our counterfactual generator computes the categorical distribution as follows,
\begin{equation}
\begin{aligned}
    \tilde{p}_{tk} = & P(Y_t=k|\boldsymbol{X}=\boldsymbol{x}, \boldsymbol{Z}=\boldsymbol{\tilde{z}}, \boldsymbol{Y}_{<t}=\boldsymbol{\tilde{y}}_{<t}),\\
    \tilde{y}_t =  & \underset{k=1, \ldots,|V|}{\arg \max}(\log \tilde{p}_{tk} + u_{tk}),
\end{aligned}
\end{equation}
where $\boldsymbol{\tilde{z}}$ is the predicted reply perspective, $\boldsymbol{\tilde{y}}_{<t}$ is the token sequence generated in the previous step.

Overall, counterfactual generation via perspective transition can be used as an effective data augmentation method for open-domain dialogues to augment responses with wider semantic coverage.
We show this method in Algorithm~\ref{sec:alg}.
The algorithm takes an observed sample $(\boldsymbol{x}, \boldsymbol{y})$ as an input and loop through every keyword of $\boldsymbol{x}$ as a different focus $\boldsymbol{\tilde{c}}$.
For each $\boldsymbol{\tilde{c}}$, to sample multiple corresponding reply perspectives, we equally divide the candidate set $\mathcal{N}(\boldsymbol{\tilde{c}})$ into $K$ sub-sets $\{\mathcal{N}_1(\boldsymbol{\tilde{c}}), \dots, \mathcal{N}_K(\boldsymbol{\tilde{c}})\}$ for nested loop.
At each iteration it predicts a different $\boldsymbol{\tilde{z}}$ for perspective transition to output a counterfactual sample $(\boldsymbol{x}, \boldsymbol{\tilde{y}})$.

\subsection{Model Training}
\label{method:training}
CAPT relies on the reply perspective predictor and the counterfactual generator, which greatly influence the quality of augmentation. 
Inspired by~\citet{yang-etal-2020-generative,schick-schutze-2021-generating}, we choose a pre-trained encoder-decoder model BART~\cite{lewis-etal-2020-bart} as the backbone model.
\begin{algorithm}[t]
\caption{Data Augmentation}
\label{sec:alg}
\SetKwInput{KwInput}{Input}                
\SetKwInput{KwOutput}{Output}              
\DontPrintSemicolon
  
  \KwInput{
     $(\boldsymbol{x}, \boldsymbol{y})$: An observed sample \\
     \quad \quad \quad\,$\mathcal{C}$: All keywords $\{\boldsymbol{\tilde{c}}_1, \dots, \boldsymbol{\tilde{c}}_{|\mathcal{C}|}\}$ of $\boldsymbol{x}$ \\
     \quad \quad \quad\,$\mathcal{G}$: The shift graph \\
  }
  \KwOutput{A counterfactual sample $(\boldsymbol{x}, \boldsymbol{\tilde{y}})$}

  \SetKwFunction{FTransit}{Trans}
  
  Get the observed reply perspective $\boldsymbol{z}$; \\
  \For{$i\gets1$ \KwTo $|\mathcal{C}|$}{
    Get 1-hop neighbors $\mathcal{N}(\boldsymbol{\tilde{c}}_i)$ from $\mathcal{G}$ \;
    Remove $\boldsymbol{z}$ from $\mathcal{N}(\boldsymbol{\tilde{c}}_i)$ \\
    Equally divide $\mathcal{N}(\boldsymbol{\tilde{c}}_i)$ into $\{\mathcal{N}_1(\boldsymbol{\tilde{c}}_i), \dots, \mathcal{N}_K(\boldsymbol{\tilde{c}}_i)\}$ \;
    \For{$j\gets1$ \KwTo $K$}{
        $\boldsymbol{\tilde{y}}$ $\gets$ \FTransit{$\boldsymbol{x}$, $\boldsymbol{y}$, $\boldsymbol{z}$, $\boldsymbol{\tilde{c}}_i$, $\mathcal{N}_j(\boldsymbol{\tilde{c}}_i)$}
    }
    }

  \SetKwProg{Fn}{Function}{:}{}
  \Fn{\FTransit{$\boldsymbol{x}$, $\boldsymbol{y}$, $\boldsymbol{z}$, $\boldsymbol{\tilde{c}}$, $\mathcal{N}(\boldsymbol{\tilde{c}})$}}{
        Infer $\boldsymbol{u}$ from $P(\boldsymbol{U}|\boldsymbol{x}, \boldsymbol{y}, \boldsymbol{z})$ \\
        Predict $\boldsymbol{\tilde{z}}$ from $P(\boldsymbol{Z}|\boldsymbol{x}, \boldsymbol{\tilde{c}}, \mathcal{N}(\boldsymbol{\tilde{c}}))$ \\
        Reason $\boldsymbol{\tilde{y}}$ from $P(\boldsymbol{Y}|\boldsymbol{x}, \boldsymbol{\tilde{z}})$ under the current environment $\boldsymbol{u}$\\
        \KwRet $\boldsymbol{\tilde{y}}$
  }
\end{algorithm}
\paragraph{Reply Perspective Predictor.}
We fine-tune BART on $\mathcal{D}$ to learn $P(\boldsymbol{Z}|\boldsymbol{C}, \boldsymbol{X}, \boldsymbol{N})$.
In particular, the input is a concatenated text sequence consisting of the post $\boldsymbol{X}$, the focus $\boldsymbol{C}$, and the candidates $\boldsymbol{N}$.
The output is the predicted reply perspective $\boldsymbol{Z}$.
We maximize the objective as follows,
\begin{equation}
\label{eq:predict}
    \mathcal{L}_p = -\sum\nolimits_{t=1}^{|Z|}\log P(Z_t|[\boldsymbol{C}, \boldsymbol{X}, \boldsymbol{N}], \boldsymbol{Z}_{<t}),
\end{equation}
where the bracket $[\cdot, \cdot, \cdot]$ denotes concatenation with the token $[\text{SEP}]$. 
The candidates $\boldsymbol{N}$ are also concatenated with commas.
$\boldsymbol{Z}_{<t}$ is a prefix of the reply perspectives.
$|Z|$ denotes the length of  $\boldsymbol{Z}$.

\paragraph{Counterfactual Generator.} 
We fine-tune BART on $\mathcal{D}$ to learn $P(\boldsymbol{Y}|{\boldsymbol{X}, \boldsymbol{Z}})$.
Specifically, the generator is trained to generate the response $\boldsymbol{Y}$ with the input prompt consisting of the post $\boldsymbol{X}$ and the reply perspective $\boldsymbol{Z}$.
Similarly, we maximize the following objective:
\begin{equation}
\label{eq:generate}
    \mathcal{L}_g = -\sum\nolimits_{t=1}^{|Y|}\log P(Y_t|[\boldsymbol{X}, \boldsymbol{Z}], \boldsymbol{Y}_{<t}),
\end{equation}

\subsection{Bi-directional Perplexity Selection}
\label{method:selection}
Filtering out detrimental augmented samples can improve downstream performance~\cite{pmlr-v119-bras20a}. 
Existing methods~\cite{axelrod-etal-2011-domain,NEURIPS2020_44feb009,zhang-etal-2020-dialogue} pick out samples that the model only trained on the observed data is most confident about.
However, these models have only seen limited samples so they may not identify valid but unseen samples from the counterfactual-generated data.
Inspired by~\citet{lee-etal-2021-towards}, we leverage a large-scale dialogue pre-trained language model DialoFlow~\cite{li-etal-2021-conversations}, utilizing its powerful ability of transfer learning.
Since large-scale dialogues have been seen, it can identify valid but unseen samples like ``an expert'' via perplexity (PPL) scores.
Nonetheless, the resulting samples might contain samples with generic responses.
Inspired by~\citet{li-etal-2016-diversity}, we further introduce backward PPL to rerank responses for prioritizing those valid and interesting samples.

Specifically, we independently fine-tune DialoFlow to learn $P(\boldsymbol{Y|X})$ and $P(\boldsymbol{X|Y})$ on $\mathcal{D}$ for calculating \emph{forward} and \emph{backward} PPL scores. 
Once we obtain the forward PPL scores for all samples, we find the best threshold $\eta$ that separates valid samples from invalid samples.
Inspired by~\citet{lee-etal-2021-towards}, we leverage the validation set to find the optimal single threshold parameter $\eta$, where we regard observed samples from the validation set as valid samples, and invalid samples are constructed by replacing the responses of valid samples with randomly-sampled responses.
Furthermore, we rerank the responses of each post in the valid samples via backward PPL scores.
Since the higher the backward PPL score, the more likely the response is dull~\cite{li-etal-2016-diversity},
we choose samples in order from low to high until the desired number of augmented samples are obtained.

\section{Experimental Setup}
\subsection{Settings}
The experiments are conducted on the Chinese Weibo corpus~\cite{zhang-etal-2020-dialogue}.
Specifically, the dataset $\mathcal{D}$ contains training, validation, and test sets with $300$K, $5$K, and $10$K post-response samples, respectively.
Please see Appendix~\ref{app:setting} for more details on data and method implementations.
\begin{figure*}
    \centering 
    \includegraphics[width=1\linewidth]{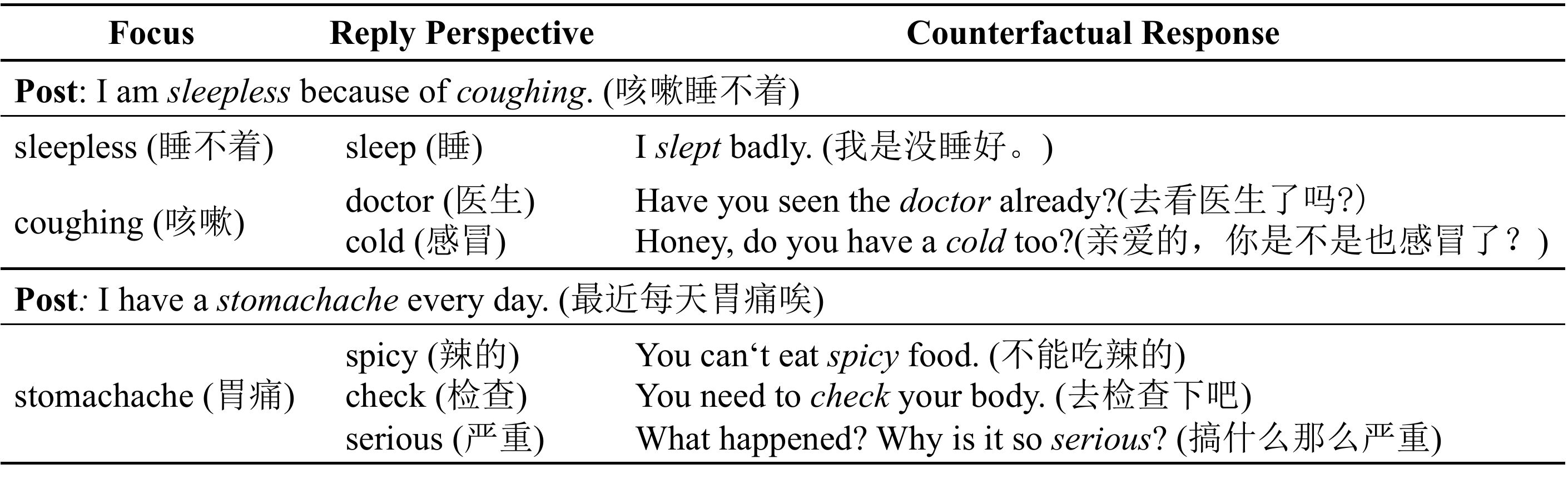}     
    \caption{Real cases showing the generation process of responses with different semantics.
    } 
    \label{fig:case} 
\end{figure*}


\subsection{Baselines}
We compare CAPT with a set of baselines:
(1) \textbf{Observed}, which only uses the observed data to fine-tune dialogue models.
(2) \textbf{Augmented}, which only uses our augmented data to fine-tune dialogue models.
(3) \textbf{Back-Trans}~\cite{sennrich-etal-2016-improving}, which back-translates responses via Google Translate.
(4) \textbf{MLM}~\cite{cai-etal-2020-data}, which fine-tunes the BERT-large model on $\mathcal{D}$ to substitute some words of responses. The substituting probability is 0.15.
(5) \textbf{DL}~\cite{zhang-etal-2020-dialogue}, which constructs post-response pairs where both post and response are retrieved from the unpaired data. Augmented dialogues are further filtered by their ranking module.
(6) \textbf{BM25}~\cite{gangal-etal-2021-improving}, which uses the BM25 algorithm to retrieve the top-k similar post to the observed post, and the corresponding response of the retrieved post is regarded as the augmented response. 
(7) \textbf{BART}~\cite{lewis-etal-2020-bart}, which fine-tunes the BART-large model that takes the post as the input to generate responses with different decode strategies, including greedy search, sampling with temperature 0.5, and top-k sampling (k=10,25).
They are denoted as \textbf{BART-greedy}, \textbf{BART-samp}, \textbf{BART-k10}, and \textbf{BART-k25}, respectively.
Augmented pairs generated by BM25 and BART are filtered by our data selection method.
\subsection{Evaluation Metrics}
\paragraph{Automatic Evaluation.}
The following metrics are used to automatically evaluate \emph{retrieval-based} models. (1) Mean Average Precision (\textbf{MAP}): the average of Average Precision (AP) over test samples. AP is the average of precision scores at the ranks where references are found; (2) $\boldsymbol{R_{10}@k}$: the percentage of references among the top-k selected responses (k=1,2,5) when given 10 candidates in total.
The following metrics are used to evaluate \emph{generation-based} models. 
(1) \textbf{BLEU}: the overlap of n-grams (n<4) between the generated response and the reference. 
(2) \textbf{Dist-n}: the ratio of unique n-grams (n=1,2) over all n-grams in the generated responses, which measures the n-gram diversity. 
As we sample 3 responses for each test post, evaluation is performed both within and among the sampled responses. 
\textbf{Intra-Dist} calculates that ratio within 
each sampled response, and \textbf{Inter-Dist} calculates that ratio among all three responses.
(3) $\textbf{BS}_f$: the F1-value of \textbf{BERTScore}~\cite{Zhang*2020BERTScore:}, which measures the semantic similarity between each 2 responses in 3 sampled responses. Lower scores imply greater semantic diversity.

We also use \textbf{Dist-n} and $\textbf{BS}_f$ to automatically evaluate the quality of augmented data, which evaluates the diversity among the generated responses.
In addition, we introduce the following metrics to evaluate the diversity with respect to the original response. 
(1) \textbf{Novelty-n}: 
the ratio of new n-grams (n=1,2) in the augmented responses.
\textbf{Intra-Novelty} similarly calculates the ratio within each augmented response, i.e., n-grams that are covered by the augmented response but not in the original response.
\textbf{Inter-Novelty} calculates the ratio within the three augmented responses.
(2) $\textbf{BS}_{fo}$: the F1-value of \textbf{BERTScore}, which measures the semantic similarity between the augmented response and its corresponding original response. 
\begin{table*}
    \small
    \centering
    \begin{tabular}{lllllllllll}
    \toprule
        \textbf{Method} &  \multicolumn{2}{c}{\textbf{Intra-Dist-1,2}}   & \multicolumn{2}{c}{\textbf{Inter-Dist-1,2}}  &  $\textbf{BS}_f$ & \multicolumn{2}{c}{\textbf{Intra-Novelty-1,2}}   & \multicolumn{2}{c}{\textbf{Inter-Novelty-1,2}} & $\textbf{BS}_{fo}$ \\ \toprule
        BART-greedy & 93.34$^\ddagger$ & 98.37$^\ddagger$  & 64.83$^\ddagger$ & 81.81$^\ddagger$ & 66.46$^\ddagger$ & 84.42$^\ddagger$ & 95.54$^\ddagger$ & 60.54$^\ddagger$ & 80.38$^\ddagger$ & 58.12$^\ddagger$ \\ 
        BART-samp & 94.14$^\ddagger$ & 98.79$^\ddagger$  & 70.85$^\ddagger$ & 89.27$^\ddagger$ & 63.60$^\ddagger$ & 84.24$^\ddagger$ & 95.99$^\ddagger$ & 65.84$^\ddagger$ & 87.74$^\ddagger$ & 58.11$^\ddagger$ \\ 
        BART-k10 & 93.08$^\ddagger$ & 98.63$^\ddagger$  & 70.60$^\ddagger$ & 90.07$^\ddagger$ & 63.15$^\ddagger$ & 85.00$^\dagger$ & 96.23$^\dagger$ & 67.36$^\ddagger$ & 89.11$^\ddagger$ & 58.08$^\ddagger$\\ 
        BART-k25 & 93.74$^\ddagger$ & 98.77$^\ddagger$ & 74.63$^\ddagger$ & 91.98$^\ddagger$ & 61.61$^\ddagger$ & 85.76  & 96.43  & 71.01$^\ddagger$ & 90.90$^\ddagger$ & 57.83$^\ddagger$ \\ 
        \textbf{CAPT} & \textbf{94.64}  & \textbf{98.90}  & \textbf{79.91}  & \textbf{94.79}  & \textbf{59.59}  & \textbf{85.84}  & \textbf{96.63}  & \textbf{74.47}  & \textbf{92.98}  & \textbf{57.31}$^\ddagger$ \\ \toprule
        Observed & 94.05  & 98.90  & - & - & - & -  & -  & -  & - & -\\ \toprule
    \end{tabular}
    \caption{Automatic evaluation on the quality of augmented data generated by different generation-based methods.
    The bottom row corresponds to the high-quality observed dialogue data in $\mathcal{D}$.
    Significance tests between CAPT and baselines are performed using t-test.
    $\dagger$ and $\ddagger$ indicate $p$-value < 0.05 and 0.01, respectively.    }
    \label{tab:auto_aug}
\end{table*}
\begin{table}
    \small
    \centering
    \begin{tabular}{llllll}
    \toprule
        \textbf{Method} &  \textbf{Flu.} & \textbf{Coh.} & \textbf{Int.} & \textbf{Rich.} \\ \toprule
        BART-greedy  & 1.921  & 1.507 & 1.222$^\ddagger$ & 0.611$^\ddagger$  \\ 
        BART-samp & 1.833$^\ddagger$ & 1.383$^\ddagger$ & 1.500$^\ddagger$ & 0.926$^\ddagger$  \\ 
        BART-k10 & 1.853$^\ddagger$ & 1.461$^\dagger$ & 1.506$^\ddagger$ & 0.983$^\ddagger$ \\ 
        BART-k25 & 1.813$^\ddagger$ & 1.333$^\ddagger$ & 1.560$^\dagger$ & 1.182$^\ddagger$  \\ 
        \textbf{CAPT} & \textbf{1.953}  & \textbf{1.653}  & \textbf{1.707}  & \textbf{1.660}  \\ \toprule
        Observed & 1.941  & 1.744  & 1.740  & - \\ \toprule
    \end{tabular}
    \caption{Manual evaluation on augmented data.
    The bottom row corresponds to the high-quality observed dialogue data in $\mathcal{D}$.
    Significance tests between CAPT and baselines are performed using t-test.
    $\dagger$ and $\ddagger$ indicate $p$-value < 0.05 and 0.01, respectively.    }
    \label{tab:manual_aug}
\end{table}

\paragraph{Manual Evaluation.}
The following metrics are used to manually evaluate the quality of augmented data and generation-based models. Three annotators
are employed to rate the samples. 
(1) \textbf{Fluency (Flu.)}: is the response fluent?
(2) \textbf{Coherence (Coh.)}: is the response serve as a valid continuation of the preceding post?
(3) \textbf{Interesting (Int.)}: is the response generic?
(4) \textbf{Richness (Rich.)}: do the three sampled responses express different semantics?
The rating scale is of 0 to 2, in which 0 means worst and 2 best.

\section{Results and Discussion}
\subsection{Evaluating Augmented Data}
We first evaluate the quality of augmented data.
Specifically, we respectively select 900K augmented post-response pairs generated by these methods, on which automatic evaluation is performed.
We further conduct manual evaluation on 600 samples, which contain 200 randomly-sampled posts and each post has 3 corresponding responses.
The inter-annotator agreement is measured via the Fleiss's kappa $\kappa$~\cite{randolph2005free}.
The $\kappa$ values for \emph{Fluency}, \emph{Coherence}, \emph{Interesting} and \emph{Richness} are 0.67 (moderate agreement), 0.46 (moderate agreement), 0.64 (moderate agreement) and 0.69 (moderate agreement), respectively.

The results are shown in Table~\ref{tab:auto_aug} and~\ref{tab:manual_aug}, which indicates that our augmented data outperforms all the baselines. 
We further observe that: 
(1) Our augmented data achieve similar scores as the observed data over all the metrics, which indicates that our augmented data is high-quality.
We present some cases of the augmented data to show the generation process of different-semantic responses in Figure~\ref{fig:case}. 
(2) Our augmented data achieve better scores of $\text{BS}_f$, $\text{BS}_{fo}$ and Richness, which indicates that CAPT can augment more responses with different semantics.
In particular, BART-samp vs. CAPT shows the effectiveness of intervention in the reply perspective.
(3) BART-k10 achieves relatively good scores on all the metrics compared to other baselines.
This indicates that the top-k sampling (k=10) is superior to the other decoding strategies.
Thus, the top-k sampling (k=10) can be used for the following generation-based models.

\subsection{Evaluating Dialogue Model}
We further evaluate the benefit of our augmented data on retrieve-based and generation-based dialogue models.
Specifically, we follow~\citet{zhang-etal-2020-dialogue} and select 300K augmented post-response samples for all methods for a fair comparison.
We conduct automatic evaluation on 5K test data and manual evaluation on 600 samples that contain 200 randomly-sampled posts with 3 generated responses.
The $\kappa$ value for \emph{Fluency}, \emph{Coherence}, \emph{Interesting} and \emph{Richness} are 0.67 (moderate agreement) are 0.71 (substantial
agreement), 0.59 (moderate agreement), 0.48 (moderate agreement) and 0.53(moderate agreement), respectively.
\begin{table}[]
    \small
    \centering
    \begin{tabular}{lllll}
    \toprule
        \textbf{Method} & \textbf{MAP} & $\mathbf{R_{10}@1}$ & $\mathbf{R_{10}@2}$ & $\mathbf{R_{10}@5}$ \\ \toprule
        Observed & 80.21$^\ddagger$ & 69.72$^\ddagger$ & 82.05$^\ddagger$ & 94.96$^\dagger$ \\ 
        Augmented & 76.67$^\ddagger$ & 65.14$^\ddagger$ & 78.16$^\ddagger$ & 92.46$^\ddagger$  \\ 
        MLM & 80.22$^\ddagger$ & 69.76$^\ddagger$ & 82.05$^\ddagger$ & 94.90$^\dagger$  \\ 
        Back-Trans & 80.26$^\ddagger$ & 69.75$^\ddagger$ & 82.21$^\ddagger$ & 94.99  \\ 
        DL & 80.47$^\ddagger$ & 70.05$^\ddagger$ & 82.41$^\dagger$ & 95.03  \\ 
        BM25 & 80.07$^\ddagger$ & 69.68$^\ddagger$ & 81.62$^\ddagger$ & 94.82$^\ddagger$  \\
        BART-greedy & 80.37$^\ddagger$ & 70.03$^\ddagger$ & 82.17$^\ddagger$ & 94.75$^\ddagger$  \\ 
        BART-samp & 80.42$^\ddagger$ & 70.17$^\ddagger$ & 82.03$^\ddagger$ & 94.88$^\ddagger$  \\ 
        BART-k10 & 80.38$^\ddagger$ & 70.06$^\ddagger$ & 82.15$^\ddagger$ & 94.79$^\ddagger$  \\ 
        BART-k25 & 80.53$^\ddagger$ & 70.30$^\ddagger$ & 82.21$^\ddagger$ & 94.91$^\dagger$  \\ 
         \toprule
        \textbf{CAPT} & \textbf{81.08} & \textbf{71.08} & \textbf{82.86} & \textbf{95.14} \\ \bottomrule
    \end{tabular}
    \caption{Automatic evaluation on different data augmentation methods for retrieve-based models. We repeatedly experiment 10 times with different seeds and report the averaged scores.
    $\dagger$ and $\ddagger$ indicate that the improvement of CAPT is significant at the level of 0.05 and 0.01 respectively (significance tests via t-test).
    }
    
    \label{tab:retrieve}
\end{table}
\begin{table*}
    \small
    \centering
    \begin{tabular}{lllllllllll}
    \toprule
        \textbf{Method} & \textbf{BLEU} & \multicolumn{2}{c}{\textbf{Intra-Dist-1,2}} & \multicolumn{2}{c}{\textbf{Inter-Dist-1,2}} & $\textbf{BS}_f$ & \textbf{Flu.} & \textbf{Coh.} & \textbf{Int.} & \textbf{Rich.} \\ \toprule
        Observed & 2.22 & 91.11$^\ddagger$ & 98.21$^\ddagger$ & 73.83$^\ddagger$ & 93.18$^\dagger$ & 60.54$^\ddagger$ & 1.806$^\ddagger$ & 1.377$^\ddagger$ & 1.645 & 1.075$^\ddagger$  \\ 
        Augmented & 1.85 & 92.29$^\ddagger$ & 98.16$^\ddagger$ & 77.86 & 93.28 & 59.76 & 1.848 & 1.363$^\ddagger$ & 1.652 & 1.320  \\ 
        MLM & 2.16 & 91.19$^\ddagger$ & 98.25$^\ddagger$ & 74.41$^\ddagger$ & 93.37 & 60.50$^\ddagger$ & 1.813$^\dagger$ & 1.438 & 1.653 & 1.095$^\ddagger$  \\ 
        Back-Trans & 2.21 & 91.26$^\ddagger$ & 98.26$^\ddagger$ & 74.66$^\ddagger$ & 93.49 & 60.45$^\ddagger$ & 1.791$^\ddagger$ & 1.443 & 1.657 & 1.115$^\ddagger$  \\ 
        DL & 2.23 & 92.09$^\ddagger$ & 98.35$^\ddagger$ & 75.02$^\ddagger$ & 93.42 & 60.35$^\ddagger$ & 1.823$^\dagger$ & 1.462 & 1.665 & 1.135$^\ddagger$  \\ 
        BM25 & 1.68 & 91.55$^\ddagger$ & 98.14$^\ddagger$ & 76.51$^\ddagger$ & 92.02$^\ddagger$ & 60.17$^\dagger$ & 1.803$^\ddagger$ & 1.155$^\ddagger$ & 1.650 & 1.185$^\dagger$  \\ 
        BART-greedy & \textbf{3.54} & 91.54$^\ddagger$ & 98.02$^\ddagger$ & 64.79$^\ddagger$ & 80.87$^\ddagger$ & 67.18$^\ddagger$ & 1.841 & 1.453 & 1.508$^\ddagger$ & 0.895$^\ddagger$  \\ 
        BART-samp & 2.86 & 92.12$^\ddagger$ & 98.42$^\ddagger$ & 69.81$^\ddagger$ & 88.91$^\ddagger$ & 63.51$^\ddagger$ & 1.822$^\dagger$ & 1.448 & 1.582$^\ddagger$ & 0.910$^\ddagger$  \\ 
        BART-k10 & 2.72 & 91.71$^\ddagger$ & 98.53 & 70.51$^\ddagger$ & 90.02$^\ddagger$ & 63.45$^\ddagger$ & 1.835 & 1.480 & 1.584$^\ddagger$ & 0.925$^\ddagger$  \\ 
        BART-k25 & 2.70 & 91.93$^\ddagger$ & 98.45$^\ddagger$ & 71.29$^\ddagger$ & 90.46$^\ddagger$ & 62.81$^\ddagger$ & 1.812$^\dagger$ & 1.425$^\dagger$ & 1.623$^\dagger$ & 0.935$^\ddagger$  \\ \toprule
        \textbf{CAPT} & 2.11 & \textbf{93.39} & \textbf{98.67} & \textbf{78.03} & \textbf{93.62} & \textbf{59.64} & \textbf{1.867} & \textbf{1.492} & \textbf{1.677} & \textbf{1.355} \\ \bottomrule
    \end{tabular}
    \caption{Automatic and manual evaluation on different data augmented methods for generation-based dialogue models.
    Significance tests between CAPT and baselines were performed using t-test, where booststrap resampling~\cite{koehn-2004-statistical} was applied for automatic evaluation. $\dagger$ and $\ddagger$ indicate $p$-value < 0.05 and 0.01, respectively.}
    \label{tab:generation}
\end{table*}

The results on retrieve-based and generation-based models are respectively shown in Table~\ref{tab:retrieve} and~\ref{tab:generation},
which indicates that CAPT outperforms all the baselines on almost all the metrics for both dialogue models.
This confirms the effectiveness of augmenting valid responses with different semantics.
We can further observe that: 
(1) CAPT achieves higher scores for almost all the metrics compared to other BART-based methods, 
especially BART-samp.
This demonstrates that intervention in the reply perspective is effective for improving the performance of dialogue models.
(2) CAPT achieves higher $\text{BS}_f$ and Richness ratings but a relatively lower BLEU score. 
We speculate that augmenting more semantically different samples enables dialogue models to generate more responses that differ from references.
\begin{figure}[t!]
    \centering  
    \subfigure[Retrieve-based Models]{
        \label{Fig.map}
        \includegraphics[width=0.47\linewidth]{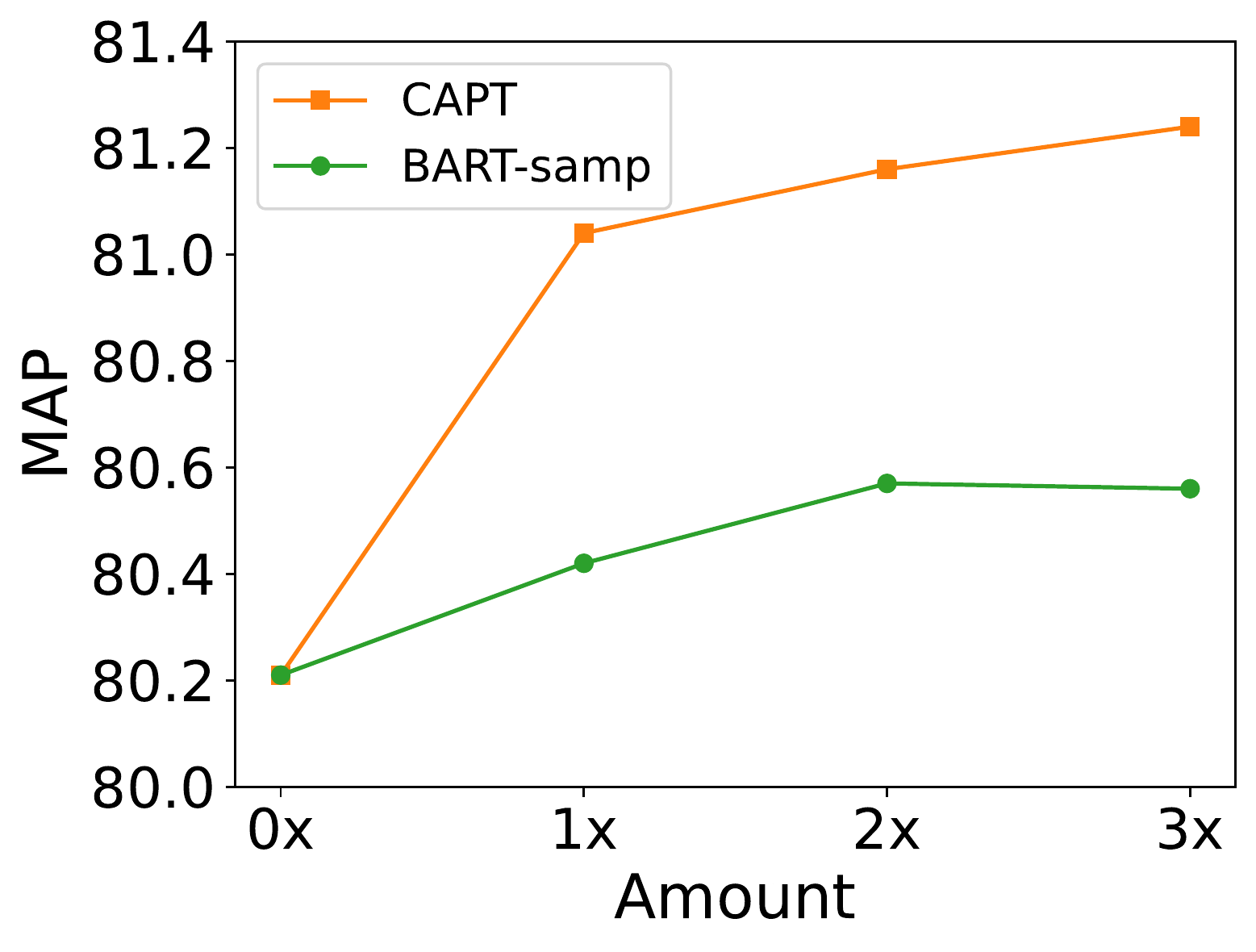}
        }
    \subfigure[Generation-based Models]{
        \label{Fig.bertscore}
        \includegraphics[width=0.45\linewidth]{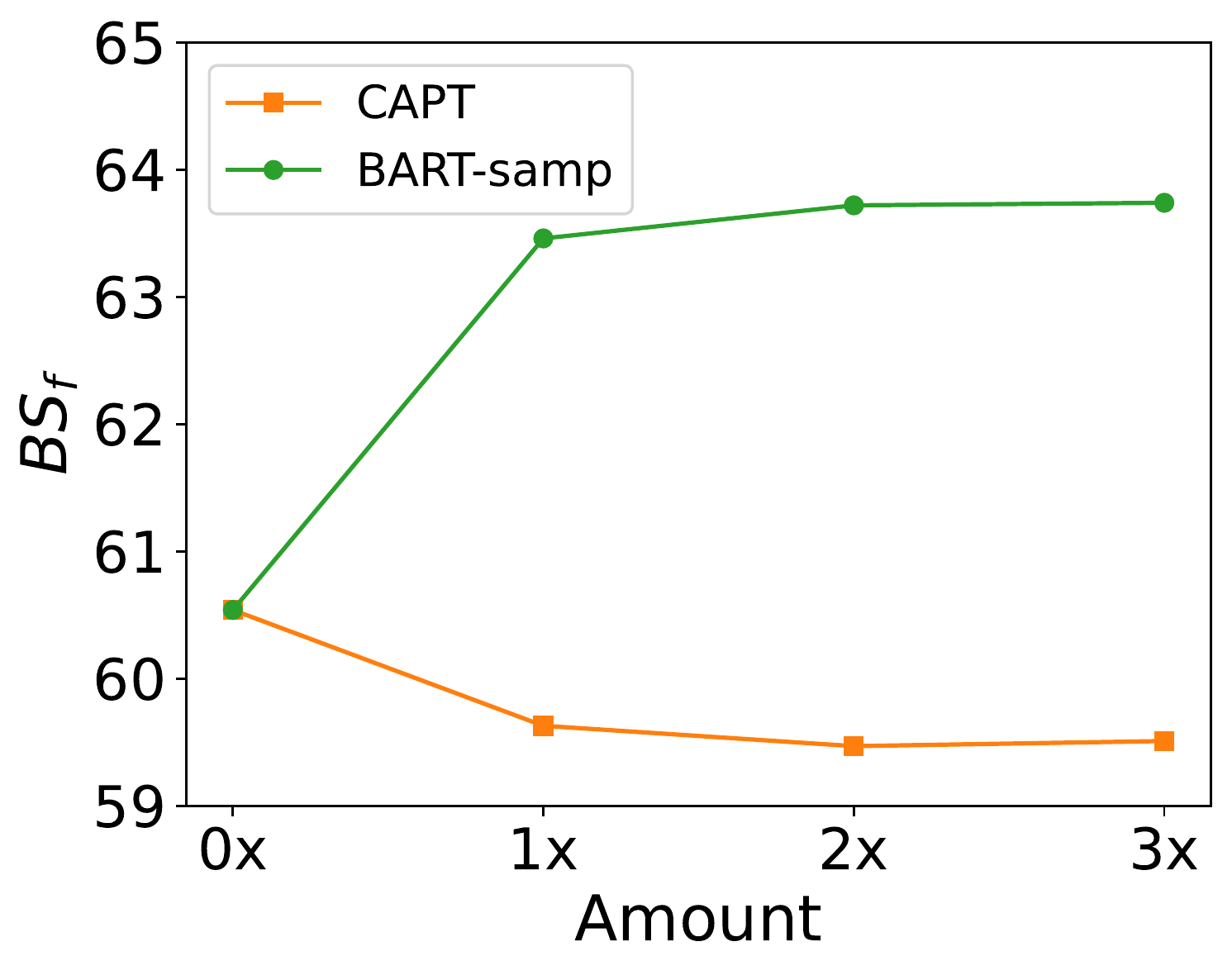}
        }
    \caption{Performance changes on retrieve-based and generation-based dialogue models respectively by providing different amounts of augmented data generated by CAPT and BART-sampling. We use MAP and $\text{BS}_f$ metrics to evaluate corresponding models. } 
    \label{Fig.amount}
\end{figure}
\begin{table}
    \small
    \centering
    \begin{tabular}{lllll}
    \toprule
        \textbf{Method} & \textbf{MAP} & $\mathbf{R_{10}@1}$ & $\mathbf{R_{10}@2}$ & $\mathbf{R_{10}@5}$ \\ \toprule
        CAPT & 81.08 & 71.08 & 82.86 & 95.14 \\ 
        -Predictor & 80.63$^\ddagger$ & 70.33$^\ddagger$ & 82.65 & 94.96$^\dagger$ \\ 
        -Candidate & 80.22$^\ddagger$ & 69.90$^\ddagger$ & 82.01$^\ddagger$ & 94.62$^\ddagger$ \\ 
        -Selection & 80.41$^\ddagger$ & 69.92$^\ddagger$ & 82.44$^\dagger$ & 95.08  \\ 
        -Dial PLM & 80.52$^\ddagger$ & 70.19$^\ddagger$ & 82.39$^\dagger$ & 94.98$^\dagger$  \\ 
        -Back PPL & 80.68$^\ddagger$ & 70.41$^\ddagger$ & 82.51$^\dagger$ & 95.07 \\ 
        -Gumbel & 80.83$^\dagger$ & 70.62$^\dagger$ & 82.76 & 95.02 \\\bottomrule
    \end{tabular}
    \caption{Ablation study on different components of CAPT on retrieve-based dialogue models.
    We repeatedly experiment 10 times with different seeds and report the averaged scores.
    $\dagger$ and $\ddagger$ indicate that the performance drop is significant at the level of 0.05 and 0.01 respectively (significance tests via t-test).}
    \label{tab:ablation}
\end{table}

\subsection{Further Discussion}
Further, we also investigate the impact of the amount of augmented responses and the effect of each component of CAPT.
\paragraph{The Impact of Amount.}
We select 0x, 1x, 2x, 3x the amount of training samples to assess the impact of providing more responses and compare CAPT with the baseline, i.e., BART-samp.
Note that 3x represents that 3*300K augmented post-response samples are selected.
Considering that samples selected in order have different interesting degrees, we eliminate the impact of interesting by uniformly selecting 900K augmented samples and randomly select from them.
The results are shown in Figure~\ref{Fig.amount}.
We can observe that: 
(1) The MAP score on BART-samp reaches a peak at 2x and drops afterward, and $\text{BS}_f$ keeps increasing from 0x to 3x augmentation.
We speculate that BART-samp only outputs alternative expressions with diversified words, which have limited semantic differences.
Augmentation of similar samples at high amounts would negatively affect training.
(2) However, the MAP score on CAPT keeps increasing and $\text{BS}_f$ does not increase.
This indicates that CAPT can augment responses with different semantics, and providing more semantically different responses can further improve the performance of downstream tasks.

\paragraph{Ablation Study.}
We perform the following ablation tests to validate the effect of each component: 
(1) Randomly choose a keyword from candidates as the reply perspective without the prediction step (-Predictor);
(2) Only take the post and the focus as the input to the predictor without 1-hop neighbors as candidates (-Candidate);
(3) Do not filter out the augmented data via data selection (-Selection);
(4) Leverage a general pre-trained language model GPT2, which does not see enough dialogue samples, to replace the dialogue pre-trained language model DialoFlow (-Dial PLM);
(5) Only use the forward PPL scores to filter out invalid samples without ranking via the backward PPL scores (-Back PPL).
(6) Generate responses not under the current environment, i.e, without the posterior Gumbel noises (-Gumbel). 
The results are shown in Table~\ref{tab:ablation}.
We observe that ablating each component brings varying degrees of performance drop.
This demonstrates the necessity of designing all these components. 


\section{Related Work}


\paragraph{Data Augmentation.}
Data augmentation has been widely used in various NLP tasks and surveyed by ~\citet{shorten2019survey,ijcai2021-631,feng-etal-2021-survey,ni2021recent,Chen2021AnES}.
Overall, data augmentation methods either add slightly modified copies of existing data or create synthetic data.
Some work propose to use heuristic rules~\cite{du-black-2018-data} or paraphrasing-based methods~\cite{niu-bansal-2019-automatically,Li_Qiu_Tang_Chen_Zhao_Yan_2019,cai-etal-2020-data,zhang-etal-2020-dialogue,xie2022targetside,cao-etal-2022-model}.
Another line of work~\cite{chang-etal-2021-neural,yang-etal-2020-generative,schick-schutze-2021-generating,wang2021towards,DBLP:journals/corr/abs-2202-13047} is exploiting large-scale pre-trained language models for data augmentation.
However, these existing methods 
do not focus on creating semantically different responses.

\paragraph{Semantically Different Augmentation.}
\citet{gangal-etal-2021-improving} utilizes knowledge sources, including COMET~\cite{bosselut-etal-2019-comet} and corpus retrieval~\cite{conf/trec/RobertsonWJHG94} to augment semantically diverse references for dialogue evaluation.
Both methods only pre-define limited augmented perspectives.
In contrast, CAPT obtains richer reply perspectives by building a shift graph.

\paragraph{Counterfactual Inference.}
Our work is based on counterfactual inference~\cite{pearl2000models}, which has shown promising results in various NLP tasks, including question answering~\cite{paranjape-etal-2022-retrieval,yu-etal-2021-cosy}, machine translation~\cite{liu-etal-2021-counterfactual} and story generation~\cite{qin-etal-2019-counterfactual,Hao_Pang_Lan_Wang_Guo_Cheng_2021,Chen_Gan_Cheng_Zhou_Xiao_Li_2022}.
In particular, \citet{zhu-etal-2020-counterfactual} uses counterfactual inference for response generation, which explores potential responses via counterfactual off-policy training.
However, CAPT focuses on \emph{counterfactual data augmentation}, which can be used to improve the performance of multiple downstream tasks.

\paragraph{Graph Construction.}
Some researches~\cite{xu-etal-2020-conversational,zou-etal-2021-thinking} also build a graph to manage concept shifts for response generation, which aims to form a more coherent and controllable dialogue.
In contrast, CAPT builds a shift graph to predict valid substituted reply perspectives, which are used to augment responses with different semantics.
Due to the different purposes of use, our graph construction is different from these existing works.


\section{Conclusion}
This paper presents a counterfactual data augmentation method, CAPT, to augment more responses with different semantics for a given dialogue history.
Specifically, CAPT employs counterfactual inference to generate counterfactual responses by intervening in the observed reply perspective, which replaces with different reply perspectives for generating semantically different responses.
Experimental results show that CAPT can augment high-quality responses with different semantics, which can be further used to improve the performance of downstream tasks.
In future work, we plan to explore an appropriate training strategy for further preventing dialogue models from being affected by noises in our augmented data, and extend CAPT on multi-turn dialogues.
We hope that CAPT will encourage future research for other generation tasks.

\section*{Limitations}

CAPT works well in scenarios with a certain amount of observed data. A small amount of observed data would lead to a small-scale shift graph.
Thus, it is difficult to provide enough candidates to pick out more valid reply perspectives, and then augment sufficient valid post-response samples.
In addition, CAPT may be more suitable for open-domain dialogue augmentation in some languages that require good-quality keyword extraction methods and pre-trained models for that language. e.g., Chinese and English. 
When transferred to different languages, e.g., English, the modifications are required as follows: (1) use the English-version keyword extraction method and keyword/sentence encoder when building the graph; (2) use the English-version pre-trained model as the backbone model for the reply perspective predictor and the counterfactual generator. 

\section*{Ethics Statement}
In this work, we employ three annotators to manually evaluate the quality of augmented data and generation-based dialogue models.
We pay $\$0.2$ to each annotator for each sample.

\section*{Acknowledgement}
We sincerely thank the anonymous reviewers for
their thorough reviewing and valuable suggestions.

\bibliography{emnlp2022}
\bibliographystyle{acl_natbib}

\clearpage
\appendix

\section{Task Definitions}
\label{app:definition}
\paragraph{Response Selection.}
Given a dataset $\mathcal{D}=\{(\boldsymbol{x}^i, \boldsymbol{y}^i, l^i)_{i=1}^N\}$, the retrieval-based dialogue model learns a matching function to correctly identify the positive response from a set of negative responses.
Specifically, the matching function $P_\theta(l^i|\boldsymbol{x}^i, \boldsymbol{y}^i)$ predicts whether the response $\boldsymbol{y}^i$ matches the dialogue history $\boldsymbol{x}^i$.
$l^i \in \{0, 1\}$ denotes a matching label, which indicates that $\boldsymbol{y}^i$ is a proper response for $\boldsymbol{x}^i$ if $l^i=1$, otherwise $l^i=0$.
The model parameters $\theta$ can be learned by minimizing the loss function that is formulated as
\begin{equation}
\begin{aligned}
    \mathcal{L}_{sel} = & -\sum\nolimits_{i=1}^N[l^i\log P_\theta(l^i=1|\boldsymbol{x}^i, \boldsymbol{y}^i) \\
    & + (1-l^i)\log P_\theta(l^i=0|\boldsymbol{x}^i, \boldsymbol{y}^i)]
\end{aligned}
\end{equation}
Generally, the training negative responses are randomly selected from the dataset $\mathcal{D}$.
\paragraph{Response Generation.}
Given a dataset $\mathcal{D}=\{(\boldsymbol{x}^i, \boldsymbol{y}^i)_{i=1}^N\}$, the generation-based dialogue model learns to model the distribution $\mathcal{P}_\phi(\boldsymbol{y}^i|\boldsymbol{x}^i)$ of the response $\boldsymbol{y}^i$ given the dialogue history $\boldsymbol{x}^i$.
The model parameters $\phi$ can be learned by minimizing the following loss:
\begin{equation}
    \mathcal{L}_{gen} = -\sum\nolimits_{i=1}^{N}\log P_\phi(\boldsymbol{y}^i|\boldsymbol{x}^i)
\end{equation}

However, a dialogue dataset that admits multiple semantically different responses for each dialogue history is usually expensive to collect, as it requires annotators to write a large variety of valid responses. 
Although such a dataset can be crawled from social networks, it will contain many noisy and meaningless responses.
It is also expensive to pick out sufficient high-quality dialogues that meet requirements.
Thus, counterfactual data augmentation aims to further augment  different-semantic responses $\boldsymbol{\tilde{y}}^i$ for  $\boldsymbol{x}^i$ in $\mathcal{D}$ without manually collecting new data.
In the following sections, we will omit the superscript $i$ for simplicity.

\section{Experimental Details}
\label{app:setting}
\subsection{Data}
The experiments are conducted on the Chinese Weibo corpus~\cite{zhang-etal-2020-dialogue}. Specifically, the dataset $\mathcal{D}$ contains training, validation, and test sets with $300$K, $5$K, and $10$K post-response samples, respectively.
To build the shift graph, we apply YAKE~\cite{campos2020yake} that relies on the statistical features of the text to automatically extract the most important keywords of each utterance in the training data.
Keywords are limited to nouns, adjectives and verbs.
The number of keyword vertices and edges are $77,439$ and $202,266$ respectively.
Furthermore, we randomly sample 200 post-response samples and employ three human annotators to evaluate the appropriateness of both keywords of focus and reply perspective. 
About $86\%$ keyword pairs are accepted by the annotators.
The average number of candidate keywords at training and augmentation times are 102 and 124 respectively.
After achieving augmented data, we similarly evaluate whether the responses share similar core semantics with the given reply perspectives.
About $96.5\%$ responses are accepted by the annotators.

\subsection{Implementation Details}
\paragraph{CAPT.}
For graph construction, we pursue bert-as-service~\cite{xiao2018bertservice} to achieve the embedding by mapping a variable-length text sequence to a fixed-length vector.
Our predictor and generator are independently fine-tuned on the BART-large model~\cite{shao2021cpt} using the loss in Eq.~\ref{eq:predict} and~\ref{eq:generate} for ten epochs, with the batch size of 64, the learning rate of 1e-5.
The other hyper-parameter setting follows that of~\citet{shao2021cpt}.
The maximum sequence length is set to 512.
We thus limit the maximum candidate size of our predictor to 100.
If the candidate size is greater than 100, we randomly sample 100 candidates.
We then filter out those samples whose candidate size is less than 5.
For data selection, we implement the score functions by fine-tuning the pre-trained DialoFlow~\cite{li-etal-2021-conversations} model with $\mathcal{D}$ for two epochs, with the batch size of 64 and the learning rate of 1e-5.
The best threshold $\eta$ is 10.

At augmentation time, we also limit the range of the candidate size from $5$ to $100$. 
Thus, we divide the whole candidate set into $K$ sub-sets and set the candidate size of each sub-set $N_{\tilde{c}} = \max(\min(\frac{|\mathcal{N}(\boldsymbol{\tilde{c}})|}{K}, 100), 5)$, where $K$ is initialized by 20.
We further update $K=\frac{|\mathcal{N}(\boldsymbol{\tilde{c}})|}{N_{\tilde{c}}}$.
The predictor outputs reply perspectives with greedy search.
The generator samples counterfactual responses from posterior Gumbel noises, the temperature is set to 0.5.

\paragraph{Retrieve-based Model.}
The retrieve-based model is built by fine-tuning the pre-trained BERT-base~\cite{devlin-etal-2019-bert}  for two epochs, with the learning rate of 2e-5, the batch size of 64, and the max sequence length of 512.
we adopt the last checkpoint for evaluation.

\paragraph{Generation-based Model.}
The generation-based model is built by fine-tuning the pre-trained BART-large~\cite{shao2021cpt} for five epochs, with the learning rate of 1e-5, the batch size of 64, and the max sequence length of 512.
At inference time, we use the top-k sampling (k=10), and the maximum decoded length is set to 50.
we adopt the last checkpoint for evaluation.

\paragraph{Training and Evaluation.}
We train retrieve-based dialogue models with 4 GPUs, generation-based models with 8 GPUs, the reply perspective predictor with 8 GPUs, and the counterfactual generator with 8 GPUs. We use Nvidia Tesla V100 GPUs.
The training time for retrieve-based models, generation-based models, the reply perspective predictor, and the counterfactual generator is approximately 2h, 4h, 4h and 5h, respectively.
At augmentation time, it takes 55min to predict reply perspectives and 1h to generate counterfactual responses for all augmented samples.
When calculating the forward and backward PPL scores, it takes 40min respectively.

\end{document}